\documentclass[sigconf]{acmart}
\usepackage{graphicx}
\usepackage{booktabs}
\usepackage{multirow}
\usepackage{makecell}
\usepackage{float}
\usepackage{balance}

\AtBeginDocument{%
  }

\setcopyright{acmlicensed}
\copyrightyear{2024}
\acmYear{2024}
\acmDOI{XXXXXXX.XXXXXXX}

\acmConference[]{}{}{}
\acmISBN{978-1-4503-XXXX-X/18/06}

\acmSubmissionID{papers784s1}

\renewcommand\footnotetextcopyrightpermission[1]{}
\newcommand{\Skip}[1] {}
\newcommand{\et}{et al.}

\newcommand{\refFig}[1]{Figure \ref{#1}}
\newcommand{\refEq}[1]{Equation (\ref{#1})}

\newcommand{\refSec}[1]{Section \ref{#1}}
\newcommand{\refTab}[1]{Table \ref{#1}}

\begin{document}

\title{HR Human: Modeling Human Avatars with Triangular Mesh and High-Resolution Textures from Videos}

\author{Qifeng Chen}
\affiliation{
  \city{Hangzhou}
  \country{China}
} 
\email{cqf7419@gmail.com}

\author{Rengan Xie}
\affiliation{%
  \city{Hang Zhou}
  \country{China}
  }
\email{rgxie@zju.edu.cn}

\author{Kai Huang}
\affiliation{%
  \city{Hang Zhou}
  \country{China}
  }
\email{huangkai21@mails.ucas.ac.cn}

\author{Qi Wang}
\affiliation{%
  \city{Hang Zhou}
  \country{China}
  }
\email{wqnina1995@gmail.com}

\author{Wenting Zheng}
\affiliation{%
  \city{Hang Zhou}
  \country{China}
  }
\email{wtzheng@cad.zju.edu.cn}

\author{Rong Li}
\affiliation{%
  \city{Hang Zhou}
  \country{China}
  }
\email{skillzero.lee@gmail.com}

\author{Yuchi Huo}
\affiliation{%
  \city{Hang Zhou}
  \country{China}
  }
\email{huo.yuchi.sc@gmail.com}

\renewcommand{\shortauthors}{author et al.}

\begin{abstract}
Recently, implicit neural representation has been widely used to generate animatable human avatars. However, the materials and geometry of those representations are coupled in the neural network and hard to edit, which hinders their application in traditional graphics engines. We present a framework for acquiring human avatars that are attached with high-resolution physically-based material textures and triangular mesh from monocular video. Our method introduces a novel information fusion strategy to combine the information from the monocular video and synthesize virtual multi-view images to tackle the sparsity of the input view. We reconstruct humans as deformable neural implicit surfaces and extract triangle mesh in a well-behaved pose as the initial mesh of the next stage. In addition, we introduce an approach to correct the bias for the boundary and size of the coarse mesh extracted. Finally, we adapt prior knowledge of the latent diffusion model at super-resolution in multi-view to distill the decomposed texture.  Experiments show that our approach outperforms previous representations in terms of high fidelity, and this explicit result supports deployment on common renderers.
\end{abstract}



\begin{teaserfigure}
  \includegraphics[width=\linewidth]{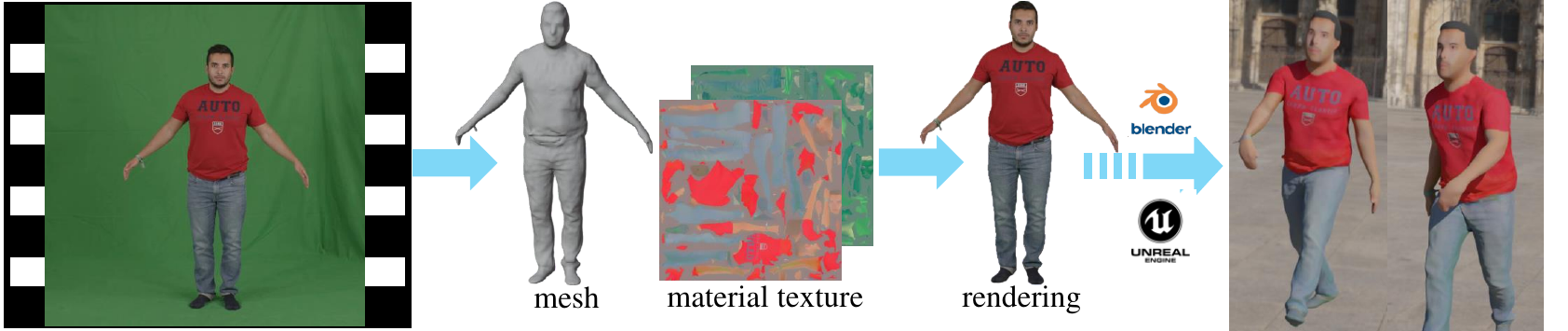}
  \caption{Given a monocular video of a performer, our method reconstructs a digital avatar equipped with high-quality triangular mesh and high-resolution corresponding PBR material textures. The result is compatible with standard graphics engines and can be edited.}
  \label{fig:teaser}
\end{teaserfigure}


\maketitle
\section{Introduction}
\label{sec:intro}
 Digital avatars have been widely used across various applications, such as in the metaverse and film production. However, producing a high-fidelity digital avatar equipped with complex attributes, including geometry, texture parameters, and material baking, requires complex pipelines and expensive equipment~\cite{scanimate, modeling, dressing, hq_sfv, A-nerf}, which limits the use of ordinary creators.
Recently, research on neural implicit representation ~\cite{nerf,deepsdf,neus, volsdf, muller2022instant} has shown impressive results in multi-view reconstruction. Advances in neural volume rendering have soon fueled various exciting works on recovering digital avatars. For the implicit animatable human reconstruction, recent works~\cite{anerf,neuralbody,neural_actor,humannerf,snarf} have solved the challenging task of multi-view reconstruction without the supervision of 3D information and present the inherent challenges of rendering non-rigid bodies and skins under dynamic motion. At the same time, inspired by neural reflectance decomposition\cite{nerfactor, nerd}, Relighting4D \cite{relighting4d} and Relightavatar~\cite{relightavatar} have attempted to recover human avatars with decoupled geometry and materials with implicit representation. However, the implicit geometry and texture are hard to edit, and the texture produced by those methods suffers from low clarity. In addition, the digital avatars represented implicitly cannot be applied in traditional graphics engines, which hinders their application in various fields.

Obviously, explicit representations appeal to us. Nvdiffrec~\cite{nvdiffrec} is dedicated to reconstructing general static objects in explicit representation that can be deployed in traditional graphics engines with triangle meshes and corresponding spatially-varying materials properties. The physical differentiable rasterization renderer has natural advantages for learning surface texture and fast rendering. However, Nvdiffrec struggles to reconstruct geometry and texture from sparse views. Furthermore, it fails to reconstruct a human in motion from monocular video. And recently 3D Gaussian(point-based rendering)\cite{3dgs} has shown great potential in dynamic human\cite{ani-3dgs}. Although a unity compatibility plugin has been released in the community\cite{unitygs}, the obvious software limitations and lack of editing are truly concerning. In contrast, our goal is to \textbf{reconstruct mesh digital avatars, taking as input the monocular video that records a human in motion, which is more compatible with traditional graphics engines and supports relighting and editing}. 

In this work, we propose a novel framework for acquiring human avatars attached with high-resolution physically-based (PBR) material textures and triangular mesh from monocular video. We first reconstruct a human as a deformable neural implicit surface using volume rendering. In this process, the human information of frames in time sequence is integrated into the deformable neural implicit surface.  From this, we extract the coarse triangle mesh and synthesize images of the human from dense virtual cameras. The synthesized images, alongside the input view, serve as supervisory data for subsequent training. We refer to this process as an information fusion strategy that combines the information from sequential video frames to compensate for the lack of spatial multi-view information. Then, we optimize the geometry mesh, material decomposition, and lighting using a differentiable rasterization renderer in the supervision of the multi-view images.In addition, We correct the bias for the boundary and size of the coarse mesh extracted from the implicit field. Finally, we introduce to adapt prior knowledge of the pretrained latent diffusion model~\cite{ldm} at super-resolution texture to distill the high-resolution decomposed texture. The LDM model has demonstrated superior performance and generalization compared to CNN pretrained network~\cite{texsr} in various vision tasks~\cite {ldm}. It fully fits our pipeline and enrichs texture space for details representation.

In summary, our main contributions are:
\begin{itemize}
\item We propose a novel framework that enables reconstructing a digital avatar equipped with triangular mesh and corresponding PBR material texture from monocular video. The digital avatars produced by our method are compatible with standard graphics engines.
\item We propose an information fusion strategy to tackle the issue of lacking multi-view supervision in reconstructing explicit geometry and texture, which integrates information from all frames in the temporal sequence of video, transforming it into spatial supervision.
\item We propose an approach to correct the bias for the triangular mesh and introduce the latent diffusion model to conduct distillation on super-resolution PBR texture. We result in a high-resolution texture and mesh with greater clarity.  
\end{itemize}

\begin{figure*}[!h] 
\centering 
\includegraphics[width=\linewidth]{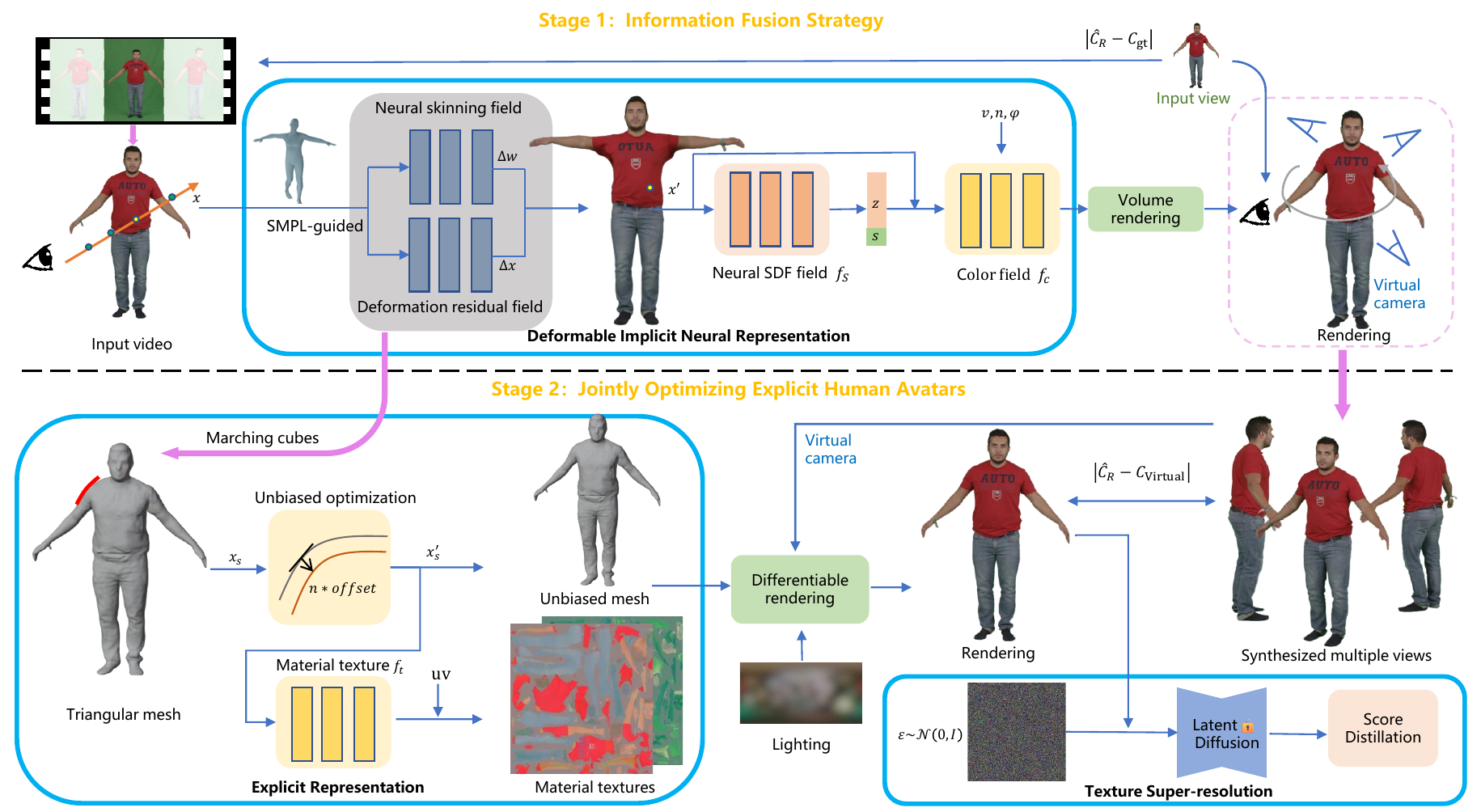}
\caption{The overview of HR Human pipeline, which takes a video frame as input to reconstruct explicit avatars with triangular mesh and PBR texture. The pipeline includes deformable neural representations (used to extract volume surfaces and enhance spare input view), explicit representations (texture and geometry are jointly optimized), and super-resolution texture modules (introduced to generate high-resolution textures). } 
\label{fig:overview} 
\end{figure*}

\section{Related Work}
\label{sec:related_work}
\subsection{Scene Reconstruction}
\label{sec:Scene Reconstruction}
Recently, neural implicit representations have achieved impressive results in 3D reconstruction~\cite{nerf,nerfw,pointNerf,maskloss,fastnerf,d-nerf}. These approaches represent a scene as a field of radiance and opacity, enabling the synthesis of photo-realistic novel viewpoints. However, directly using density-based methods for representation leads to numerous geometric artifacts. VolSDF~\cite{volsdf} and Neus~\cite{neus} proposed training the Signed Distance Function (SDF) field using volume rendering, which facilitates easy access to geometric surface normals. To further decouple the material properties, NeRD~\cite{nerd} is capable of learning geometry and spatially-varying Bidirectional Reflectance Distribution Functions parameters of objects from unconstrained environmental illumination. TensorIR~\cite{tensorir} utilizes the low-rank tensors to simultaneously estimate the geometry and material of the scene. In addition, some methods integrate deep learning with texture-based techniques to model scenes~\cite{deferred,photorealistic,neutex}. However, the neural implicit representation cannot be applied in traditional graphics engines, which hinders their application in various fields. The recent trend is point-based rendering of 3DGS~\cite{3dgs}, and along with it comes a series of improvement work~\cite{ani-3dgs,gspro}. However, point-based representation present challenges to editing and compatibility in complex graphic production. Therefore, mesh-based rendering still has its own unique advantages. DMTet~\cite{dmtet} introduces a deformable tetrahedral grid with learned mesh topology and vertex positions and utilizes coordinate-based networks to represent volumetric texturing. And Nvdiffrec~\cite{nvdiffrec} extends DMTet to 2D multi-view supervision in static scenes, jointly optimizing mesh and corresponding PBR Texture. Decomposing geometry and appearance from images contributes significantly to the progression of downstream tasks.

\subsection{Image and Texture SR}
Super-resolution is a commonly used approach to enhance detail expression in images or textures before being applied to downstream tasks of computer vision. The landscape of image super-resolution research encompasses a variety of influential works. The early super-resolution neural networks evolved from convolutional neural networks~\cite{dbsr} to GAN~\cite{bsrgan}, and later Transformer-based super-resolution~\cite{swimir} has achieved astonishing results. Super-resolution models are present in many application scenarios, such as mobisr~\cite{lee2019mobisr}, TexSR~\cite{texsr}, NeRFSR~\cite{nerfsr}. Recently, diffusion models have also shown talent in the domain of super-resolution~\cite{ldm, sr3}, diffusion models achieve highly competitive performance on various super-resolution tasks, and exploring their practicality in graphics seems valuable. In this work, we introduce to adapt prior knowledge of the pretrained latent diffusion model~\cite{ldm} at super-resolution texture to distill the high-resolution decomposed texture.

\subsection{Human Reconstruction}
On the one hand, the single-image-based human body reconstruction method with implicit functions, such as PIFu~\cite{pifu} and ECON~\cite{econ}, have demonstrated promising outcomes. However, these approaches are not fully adaptable to video inputs nor capable of generating assets that include physically-based rendering (PBR) textures. On the other hand, advances in neural volume rendering have fueled various exciting works on recovering digital avatars. There have been numerous efforts to reconstruct humans in motion using multi-view videos~\cite{neuralbody,neural_actor,A-nerf,geng2023learning,arah,snarf,humannerf}. Further, StylePeople~\cite{stylepeople} and NeuTex~\cite{neutex} introduce neural texture to restore complex texture features. And Relighting4D and Relightavatar~\cite{relightavatar,relighting4d} have attempted to recover human avatars with decoupled geometry and materials implicit representation. Ani-GS~\cite{ani-3dgs} demonstrates the potential of 3DGS in dynamic human reconstruction. However, the methods mentioned above still have a long way to go before being directly edited and illuminated on traditional graphics engines.


\section{Method}
\label{sec:method}
\refFig{fig:overview} shows the overview of our method. The framework takes as input a \textbf{monocular video} of a human to reconstruct triangular mesh and corresponding high-resolution physically-based texture, which is compatible with traditional graphics engines and supports fast \textbf{editing and relighting}. To achieve this goal, we propose an information fusion strategy to combine human information from sequential video frames, resulting in coarse geometry mesh and multi-view synthesized images. Then, we refine the human mesh and optimize the decomposition of corresponding materials texture using a differentiable rasterization renderer in the supervision of the dense and cross-view images that are synthesized by the first stage. Finally, to acquire texture with high fidelity, we adapt prior knowledge of the latent diffusion model at super-resolution in multi-view to distill the decomposed texture. Next, we will provide a more detailed introduction to the various parts of the method.

\subsection{Information Fusion Strategy}
\label{sec:implict_surface}
For monocular video, there is only a single view in each frame, and the human exhibits different poses across different video frames. Existing methods~\cite{nvdiffrec,dmtet,zhang2021physg}, which designed for reconstructing static object from mult-view supervison, are unable to reconstruct high-quality trangular mesh and physical material texture of human body from a monocular video. Therefore, we propose an information fusion strategy to extract multi-view supervision from monocular video to augment the parse input view. Information fusion strategy is composed of two primary components: firstly, optimizing a deformable neural surface of human body from video which is aimed at fuse the sequential frame information of the human body in motion; and secondly, generating pseudo multi-view images from virtual viewpoints, which act as supervison for subsequent stages.

\textbf{\emph{Reconstruct Deformable Neural Surface.}} Inspired by volsdf\cite{volsdf}, we use a SDF-based neural network $f_s$ to represent the human model in the canonical space:
\begin{equation}
\begin{aligned}
f_s:(x')\rightarrow(s(x'),z(x')),
\end{aligned}
\end{equation}
where $s(x')$ denotes the value of signed distance field (SDF) in canonical space, and $z(x')\in \mathbb{R}^{256}$ is a feature vector that represents implicit geometric information for further learning of appearance fields. The deformation from point $x$ in pose space to point $x'$ in canonical space can be divided into the sum of rigid deformation and non-rigid deformation: 
\begin{equation}
x'=\hat x+D_i(\hat x,p(i)), \hat x=T_i(x)
\end{equation}
where $D_i(\hat x,p(i))$ denotes non-rigid deformation field that is usually limited to finetune within a small range. And $\hat x$ is a preliminary result of $T_i(x)$, i.e., we add an offset to the result of rigid deformation in canonical space. $p(i)$ is the SMPL~\cite{smpl} pose parameter of the $i$-th frame. Specifically, $D_i(\hat x,p(i))$ is also implemented using a MLP $f_{Tn}:(\hat x, p(i))\rightarrow \Delta x$.
In addition, $T_i(x)$ denotes a motion field that maps point $x$ in the pose space to canonical space. And points $x$ in pose space can be projected to canonical space based on skinning weights:
\begin{equation}
T_i(x)=\sum_j^Kw_i^j(R_i^j x+t_i^j),
\end{equation}
where $R_i^j$ and $t_i^j$ denote the rotation and translation at each joint $j$, and $w_i^j$ is the blend weight for the $j$-th joint. Following peng \et~\cite{anerf}, we use the initial blend weight from SMPL to guide the rigid deformation. So we calculate the $w_i^j$ using the sum of the blend weight of SMPL and neural blend weight as:
\begin{equation}
\begin{aligned}
{w}_i^j(x) = \Delta w + \hat{w}_i^j(x),\\
\end{aligned}
\end{equation}
where $\hat{w}_i^j$ is the coarse blend weight, calculated based on the nearest points on the surface of SMPL. $\Delta w$ is the deviation used to finetune blend weights, predicted by a MLP $f_{\Delta w}:(x, \psi_i)\rightarrow \Delta w$, and $\psi_i$ is a latentcode of frame. 

As a result, the underlying surface of the human in pose space or canonical space can be easily defined as a zero-level set of $f_s$:

\begin{equation}
\begin{aligned}
S=\{{x:f(x)=0}\}
\end{aligned}
\end{equation}

Following volsdf~\cite{volsdf}, we optimize the implicit SDF field and color field of the human in canonical space using volume rendering end to end. We define the density $\sigma$ and color $c$ as:
\begin{equation}
\sigma(x')=\begin{cases}
\alpha\Bigl(1-\frac{1}{2}exp\Bigl(\frac{s(x')}{\beta}\Bigr)\Bigr)& \text{if}\ s(x')<0, \\ 
\frac{1}{2}\alpha exp\Bigl(-\frac{s(x')}{\beta}\Bigr)\Bigr)& \text{if}\ s(x') \geq   0, \\
\end{cases}
\end{equation}
\begin{equation}
c_i(x) = f_c (x',n(x'),z(x'),v(x'),\psi_i),
\end{equation}
where $\alpha, \beta > 0$ are learnable parameters and $s(x')$ is the signed distance value of point $x'$. The color field $f_c$ takes as input the sample points $x'$, normal $n(x')$, view direction $v(x')$, geometry feature code $z(x')$, latent code $\psi_i$ and predicts the color of each point in canonical space. Then the expected color $C(r)$ of a pixel along ray $r$ can be calculated using:
\begin{equation}
\begin{aligned}
\label{eq:volume integrate}
    C(r) = \sum_{n=1}^{N}(\prod_{m=1}^{n-1}(1-\alpha_m)\alpha_n c_n),
    \alpha_n = 1-exp(-\sigma_n\delta_n),
\end{aligned}
\end{equation}
where $\delta_n = t_{n+1}-t_{n}$ is the interval between sample $n$ and $n + 1$, $c_n$ is the color of sampled point along the ray.

\textbf{\emph{Generating Pseudo Multi-view Images}} 
After successfully constructing the deformable neural surface, the information from sequential video frames is effectively integrated into the neural surface, which captures global body information and can be used to synthesize images of the human from any viewpoint under multiple poses. To train an overall texture of the explicit mesh of a human in the subsequent stage, we uniformly sample 50 viewpoints around the person, aiming to cover all observable surfaces of the human body as much as possible. On the other hand, to ensure that as many areas of the human body surface are observed as possible, we select training poses that are as stretched out as possible. Even though the viewpoints are set in a single well-behaved pose, after training to acquire the explicit mesh in the subsequent stage, this human body can be animated into any pose. Then, we utilize volumetric rendering to generate images from these viewpoints. The texture information from the video is encoded into those images and optimized into the overall texture of explicit mesh in the subsequent stage. Therefore, the information fusion strategy enables recovering explicit mesh and corresponding texture from monocular video.

\subsection{Optimizing Explicit Human Avatars}
As mentioned in ~\refSec{sec:implict_surface}, we can extract a mesh from the deformable implicit surface using the marching cubes algorithm~\cite{marchingcube}. However, the mesh obtained through marching cubes tends to be coarse due to the inherent bias of the signed distance field (SDF). To address this, we propose an unbiased optimization method to refine the mesh. We then jointly optimize the physically-based material texture and the mesh using inverse rendering.

\textbf{\emph{Unbiased Optimization for Mesh.}} 
To create a triangle mesh from the neural SDF field, we first create a $256^3$ resolution 3D grid with the same size as the pose bounding box. We only select points within a certain range of space around the bone joints to query the SDF value in the MLP network, which accelerates the time required for the Marching Cubes~\cite{marchingcube}, to extract mesh in the 3D grid. We also use maximum pooling to discard small floating objects that may exist near the real human body surface. However, We observe that the neural implicit surfaces may converge in a biased range. Specifically, the Signed Distance Function (SDF) value of a well-defined surface often deviates from 0, such as ranging between $0.001$ and $0.003$, when the Marching Cubes algorithm is applied. This results in the extracted mesh not matching the human shape and being fatter than the real human body, which hinders texture optimization. We introduce a stable and easily trainable offset that works directly on the extracted explicit mesh. We further observe that the majority of the SDF bias consistently aligns with the normal of the human surface, as shown in ~\refFig{fig:fusion_bias}. Consequently, we constrain the offset along the normal direction of the vertex.
\begin{equation}
\ x'_{s} = x_{s} - f_0*n_s
\end{equation}
where $x_s$ is a set of biased vertices, $f_0$ are learnable parameters, $n_s$ are normal vector direction of vertices. $x'_{s}$ is the result vertices applied trainable offset. Specifically, we jointly optimize the bias in the early epochs of the second stage and remove it later.

\label{sec:pbr}
\textbf{\emph{Material Model.}} Inspired by Nvdiffrec~\cite{nvdiffrec}, we represent the material properties of the human surface as a physically-based material model from Disney~\cite{disney} and directly optimize it using differentiable rendering~\cite{nvdiffrast}. We use MLP to parameterize the decomposed material properties including a diffuse term $k_d$ and an isotropic and specular GGX lobe~\cite{ggx}:
\begin{equation}
    f_t:(x'_s)\rightarrow(k_d,r,m),
\end{equation}
where $k_d$ denotes albedo color, $r$ is roughness value, $m$ is metalness factor. And specular color can be defined as:
\begin{equation}
k_s = (1-m) \cdot 0.04 + m \cdot k_d,  
\end{equation}

Specifically, we parameterize the uv texture mapping for surface mesh using Xatlas~\cite{xatlas} and sample the material model on the surface to create learnable 2D textures.This is beneficial for us to continue optimizing details with high-resolution textures and make edits to the texture (\refSec{sec:sr}). 

\textbf{\emph{Physically-based Rendering.}}
In our implementation, we follow the general rendering equation~\cite{pbr}:
\begin{equation}
    L_o(x_{s}, \omega_o) = \int_\Omega f_r(x_{s}, \omega_i, \omega_o)L_i(x_{s}, \omega_i)(\omega_i \cdot n(x_{s}))d\omega_i,
\label{eq:renderingeq}
\end{equation}
where $\omega_i$ is the incident direction, $\omega_o$ is the ougoing direction, $x_{s}$ is a surface point of humans,  $f_r(x_{s}, \omega_i, \omega_o)$ is the BRDF term, $L_i(x_{s}, \omega_i)$ is the incident radiance from direction $\omega_i$, and the integration domain is the hemisphere $\Omega$ around the surface normal $n(x_{s})$ of the intersection point.

In order to fast the performance of differentiable rendering, we use split sum approximation~\cite{physically} for lighting representation. And the \refEq{eq:renderingeq} is approximated as:
\begin{equation}
\begin{aligned}
L_o(x_{s}, \omega_o) = \int_\Omega f_r(x_{s}, \omega_i, \omega_o)(\omega_i \cdot n(x_{s}))d\omega_i \\
*\int_\Omega L_i(x_{s}, \omega_i) D(\omega_i, \omega_o)(\omega_i \cdot n(x_{s}))d\omega_i
\end{aligned}
\end{equation}

where $D$ is function representing the GGX~\cite{ggx} normal distribution (NDF), the first term represents the integral of the specular BSDF with a solid white environment light, and the second term represents the integral of the incoming radiance with the specular NDF. Both of them can be pre-integrated and represented by a filtered cubemap following Karis~\cite{Karis}. Further, we employ a differentiable version of the split sum shading model to optimize the lighting represented in a learnable trainable cubemap and the material properties.

\subsection{Super-Resolution Texture.}
\label{sec:sr}
Inspired by the impressive performance of the Latent diffusion model(LDM)~\cite{ldm} in the distillation task, we introduce it to help produce super-resolution texture with more detail. We first interpolate a coarse high-resolution 2D texture mapping ($2048^2$) from the low-resolution texture mapping($512^2$) learned from RGB render loss. The coarse high-resolution 2D texture mapping and explicit mesh are utilized to render images $R$ in each view using differentiable rendering. Then, the images $R$ are fed into the LDM model as low-resolution input, and the pretrained super-resolution LDM~\cite{ldm} is used as a teacher model.
Following the score distillation, images $R$ is noised to a randomly drawn time step $t$,
\begin{equation}
    R_t = \sqrt{\bar \alpha_t}R + \sqrt{1-\bar \alpha_t}\epsilon
\end{equation}
where $\epsilon \in N(0,I)$, and $\bar \alpha_t$ is a time-dependent constant specified by diffusion model. And the score distillation loss will be calculated and gradients are propagated from rendering pixel to learnable 2d texture.
\begin{equation}
\label{eqt:sds_loss}
\nabla_x L_{SDS} = w(t)(\epsilon_\phi(R_t,t,R)-\epsilon)
\end{equation}
where $\epsilon_\phi$ is the denoising U-Net of the diffusion model, $w(t)$ is a constant multiplier that depends on $\bar \alpha_t$.

In this manner, the prior knowledge learned from extensive common datasets within the LDM model is gradually distilled into human textures, resulting in a super-resolution texture.

\section{Training}
\label{sec:loss}
\Skip{Our training schedule is divided into two stages. In the first stage, we focus on optimizing the geometry and deformation field of the human using volume rendering.}
In the first stage, to optimize the information fusion strategy without 3D supervision, we march the ray from the camera at each frame and minimize the difference between the rendered color and the ground truth color. The loss function $\mathcal{L}_1$ is defined as:
\begin{equation}
\mathcal{L}_1 =\mathcal{L}_{color} +\mathcal{L}_{eik}+\mathcal{L}_{curv}+\mathcal{L}_{offset}+\mathcal{L}_{w}
\end{equation}
where $\mathcal{L}_{color}$ is a $L_1$ loss between images. The $\mathcal{L}_{eik}$ and $\mathcal{L}_{curv}$ are the eikonal loss and curve loss applied to smooth the geometry. In addition, $\mathcal{L}_{offset}$ is a regularization term, which constrains non rigid deformation within a small range. $\mathcal{L}_{w}$ is a consistency regularization term to consistent the neural blend field.
In the second stage, we aim to refine the mesh extracted from the first stage and produce a high fidelity decomposed PBR texture of human.The loss function $\mathcal{L}_2$ consists of the following parts:
\begin{equation}
\mathcal{L}_2 = \mathcal{L}_{render} + \mathcal{L}_{bias} + \mathcal{L}_{SDS}+\mathcal{L}_{smooth} + \mathcal{L}_{light},
\end{equation}
where the $\mathcal{L}_{render}$ is the color loss of images. $\mathcal{L}_{bias}$ is applied to optimize the residual of biased surface. $\mathcal{L}_{light}$ is regularization term~\cite{nvdiffrec} designed to penalizes color shifts. $\mathcal{L}_{smooth}$ is a smooth term that smooth the texture in human surface points. Specially, $\mathcal{L}_{SDS}$ is the score distillation loss defined as \refEq{eqt:sds_loss}.\textbf{Please refer to our supplemental material for the details of each loss term, as well as the training strategy}.


\section{Experiment}
\label{sec:experiment}
\subsection{Datasets}
\textbf{\emph{Real-World Datasets.}} 
We validate our method on two real-world datasets, including ZJU-MoCap~\cite{neuralbody} and People-Snapshot~\cite{snapshot}.  ZJU-MoCap contains multiple dynamic human videos captured by a multi-camera system. People-Snapshot contains monocular videos recording humans in rotation. In addition, the approach~\cite{joo2018total} is applied to obtain the SMPL parameters within poses. We chose the most commonly used subjects to train the experiment model, including "CoreView313" and "CoreView377" from ZJU-CoMap, "M2C" and "M3C" from People-Snapshot with a monocular camera.

\textbf{\emph{Synthesized Datasets.}} 
In order to more accurately evaluate and ablate the proposed method, we follow Renderpeople\cite{Renderpeople} to capture videos under a virtual monocular camera in the blender\cite{Blender}, including "Megan", "Josh", "Brain" and "Manuel". Each human in the synthesized dataset is equipped with the reference geometry mesh and material textures. Meanwhile, synthesized datas are allowed to generate videos with more complex actions as input.

\subsection{Baseline Comparisons}
\begin{figure*}[!h] 
\centering 
\includegraphics[width=\linewidth]{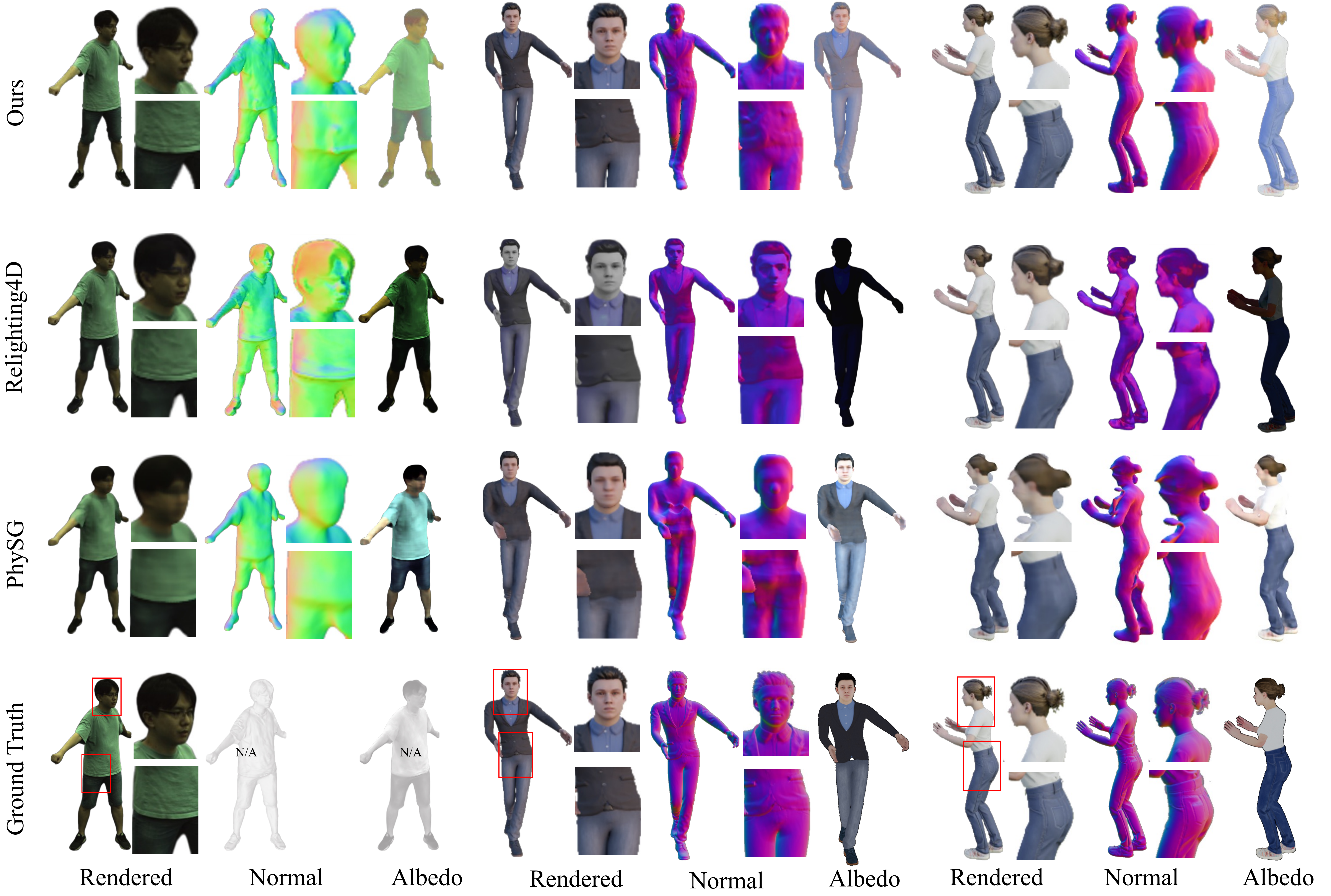}
\caption{Qualitative comparison results of comparison methods, including albedo, geometric normal, and rendered image. \Skip{Specifically, the real-world data ("ZJU313") only have the ground truth of the rendering result, and synthesized data ("Josh" and "Megan") have the reference geometry and material textures. More discussion can be found in \refSec{sec:method}.}} 
\label{fig:result} 
\end{figure*}
\textbf{\emph{Comparison Methods}} 
We compare our method with other SOTA methods that focus on reconstructing the relighting human body from videos without 3D supervision, including Relighting4D~\cite{relighting4d}, PhySG~\cite{zhang2021physg}, SDF-PDF~\cite{anerf}. Relighting4d aims to decompose the surface material and geometry, as well as the environment lighting, from the videos. PhySG focuses on recovering the geometry and material properties of static human from dense input views. Thus, we feed 120 multiview images sampled from the specific video frame as input. Even if SDF-PDF does not involve material decomposition, we also compared our method with it, which demonstrated good performance in reconstructing the dynamical human surface. Specifically, the works mentioned above reconstruct humans in the neural implicit representation. \textbf{Thus, we have added some comparison with SOTA of rendering-based human reconstruction and image-based human reconstruction in the supplementary materials.}

\textbf{\emph{Metrics}} 
Our main evaluation metrics for images include Peak Signal-to-Noise Ratio (PSNR), Structural Similarity Index Measure (SSIM)\cite{ssim}, Learned Perceptual Image Patch Similarity (LPIPS)\cite{lpips}. In addition, we follow \cite{pifu} to use 3D metrics, including Chamfer Distance(CD), Point-to-Surface Distance(P2S) and the angle degree difference between reference normals and predicted normals, which are applied to evaluate the quality of reconstructed geometry. 

\textbf{\emph{Comparison Results}}
\label{sec:method}
As shown in \refFig{fig:result}, our method outperforms the state-of-the-art works both on geometry and color appearance. Our method recovers geometry that is smoother with accurate geometric details. In addition, our method recovers convincing and clear texture details, such as the human face and the accessories of clothes, which benefit from the correct geometry and the distillate knowledge from the pretrain LDM model. \refTab{tab:comparsion} reflects the stronger capability of our method in reconstructing human body geometry and texture materials in quantity. \textbf{We show more results about materials and geometry in the supplementary materials}.

\subsection{Ablation Study}
We conduct ablation experiments from three aspects, including the effectiveness of information fusion strategy, optimization of geometric bias, and super-resolution texture distillate. Below, we provide detailed quantitative and qualitative results.

\begin{figure}[!h] 
    \centering 
    \includegraphics[width=\linewidth]{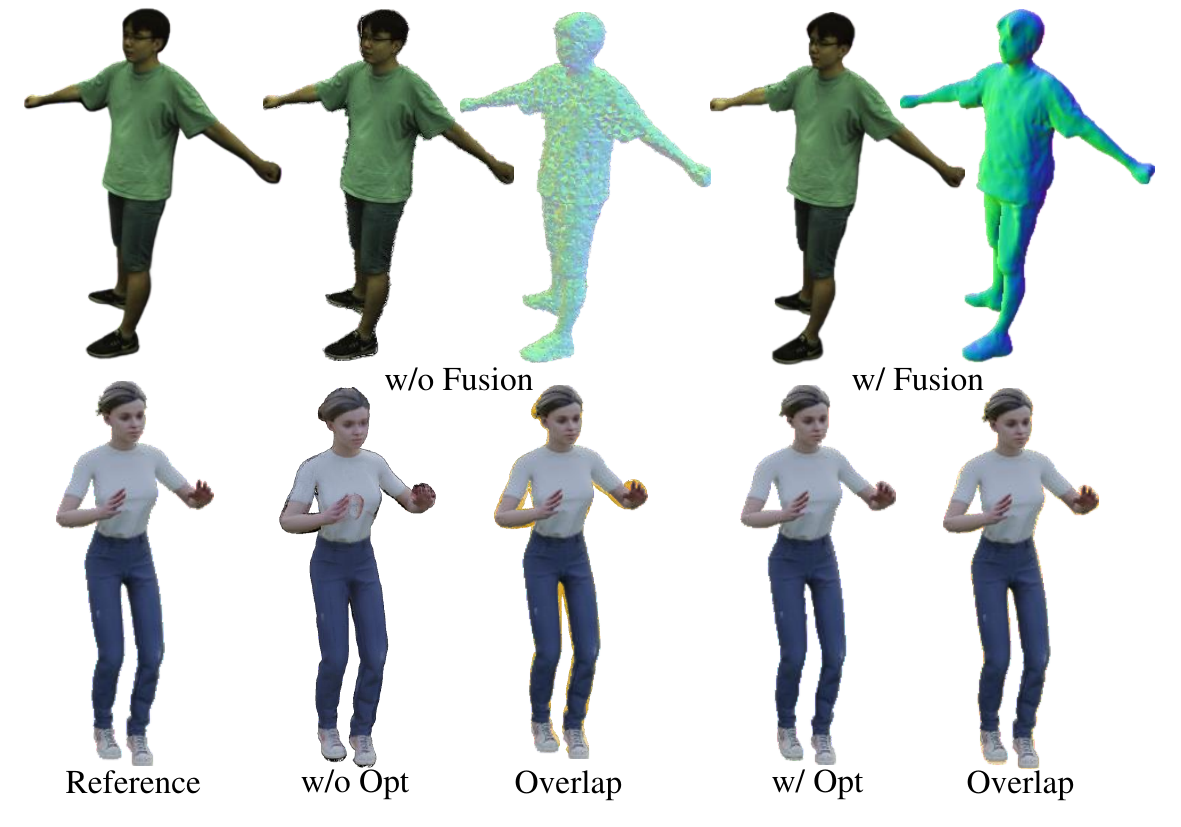}
    \caption{Qualitative comparison of the effectiveness of information fusion strategy and unbiased optimization.The highlight generated after mesh overlapping represents geometric bias.}
    \label{fig:fusion_bias} 
\end{figure}

\textbf{\emph{Information Fusion Strategy.}}
As previously mentioned, without an information fusion strategy, it would be impossible to directly train explicit mesh and PBR textures from monocular videos. Therefore, we take a step back to assess the impact of a coarse mesh on the outcomes.~\refFig{fig:fusion_bias} shows that the performance of the differentiable renderer significantly deteriorates when the coarse mesh, which is extracted from neural implicit surfaces, is removed. There are a significant amount of self-intersecting triangles in the reconstructed mesh. In contrast, our method obtained a smooth and high-quality mesh and rendering. As shown in \refTab{tab:ablation_fusion}, our method performs better in both synthesized and real-world data. Furthermore, we conducted ablation experiments to evaluate the impact of the number of synthesized virtual viewpoints on the reconstruction results. For more details, please refer to the appendix.

\begin{table}[!h]
\setlength{\tabcolsep}{1 mm}
\begin{tabular}{ccccccc}
\hline
Method      & PSNR$\uparrow$  & SSIM $\uparrow$  & LPIPS$\downarrow$ & \makecell{Normal\\Degree°$\downarrow$} &  \makecell{CD\\(cm)$\downarrow$} &  \makecell{P2S\\(cm)$\downarrow$}\\ \hline
PhySG        & 18.67 & 0.722 & 0.284 & 49.308 & - & - \\
SDF-PDF      & 24.99 & 0.919 & 0.096 & 32.712 & 0.76 & 0.70       \\
Relighting4D & 25.09 & 0.92  & 0.121 & 38.508 & 1.38 & 1.42       \\
Ours         & \textbf{27.08} & \textbf{0.94}  & \textbf{0.027} & \textbf{24.4} & \textbf{0.70} & \textbf{0.55} \\ \hline
\end{tabular}
\caption{Quantitative comparison results of various methods. The result metrics are the average of all comparison results.}
\label{tab:comparsion}
\end{table}

\begin{table}[!h]
\centering
\setlength{\tabcolsep}{0.8mm}
\begin{tabular}{cccccccc}
\hline
    DATA                  & METHOD   &  \makecell{PSNR\\$\uparrow$}   &  \makecell{SSIM\\$\uparrow$}  &  \makecell{LPIPS\\$\downarrow$}  & \makecell{Normal\\Degree°$\downarrow$} & \makecell{CD\\(cm)$\downarrow$} & \makecell{P2S\\(cm)$\downarrow$} \\ \hline
\multirow{2}{*}{ZJU313}  & w/o Fusion & 24.26  & 0.87  & 0.141  &    -   & - & -   \\
                      & w/ Fusion       & 30.68  & 0.96  & 0.028  &    - & - & -             \\ \hline
\multirow{2}{*}{Josh} & w/o Fusion  & 24.797 & 0.908 & 0.0808 & 100.25 & 0.78 &0.58     \\
                      & w/ Fusion       & 30.14  & 0.958 & 0.0264 & 24.086 &0.75 &0.58      \\ \hline
\end{tabular}
\caption{Quantitative comparison for the effectiveness of information fusion strategy on real-world data "ZJU-MoCap313" and synthesized data "Josh."}
\label{tab:ablation_fusion}
\end{table}

\textbf{\emph{Unbiased Optimization.}} The neural SDF method has always exited bias when representing zero-level surfaces. \refFig{fig:fusion_bias} shows that the mesh extracted directly from the implicit SDF field does not match with the true contour, which will affect the learning of appearance, especially at the boundary. After unbiased optimization, the mesh aligns well with the real shape, yielding better color prediction results. Table~\ref{tab:ablation_unbias and sr} further reflects the improvement of rendering results by unbiased optimized.

\begin{table}[!h]
\centering
\setlength{\tabcolsep}{5mm}
\begin{tabular}{llll}
\hline
                 & PSNR$\uparrow$  & SSIM $\uparrow$  & LPIPS $\downarrow$ \\ \hline
w/o Unbiased Opt & 23.71 & 0.89  & 0.100 \\
w/ Unbiased Opt  & 24.60 & 0.91  & 0.043 \\ \hline
NeRF SR          & 31.01 & 0.955 & 0.070 \\
w/o SR           & 30.15 & 0.910 & 0.125 \\
w/ SR            & 30.81 & 0.950 & 0.071 \\ \hline
\end{tabular}
\caption{Quantitative comparison for the effectiveness of unbiased optimization on synthesized data "Megan". And Quantitative comparison for the effectiveness of the super-resolutin texture on synthesized data "Josh".}
\label{tab:ablation_unbias and sr}
\end{table}

\begin{figure}[!h] 
\centering 
\includegraphics[width=\linewidth]{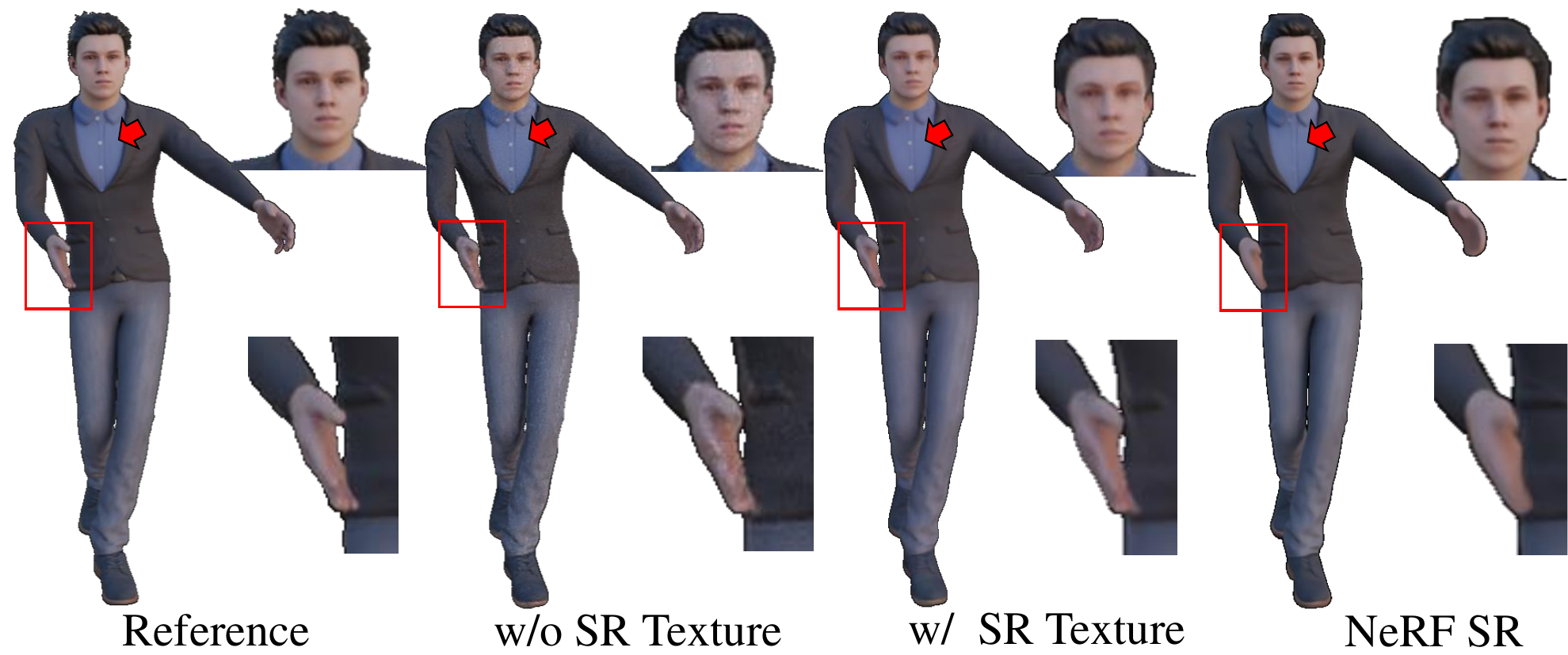}
\caption{Qualitative comparison for the effectiveness of super-resolution. The GT and the rendering results for optimized textures at $512^2$ resolution, optimized textures at $2048^2$ resolution and implicit neural field are shown from left to right separately.} 
\label{fig:ablation_sr} 
\end{figure}
\textbf{\emph{Super-Resolution Texture.}}
We design three ablation experiments to investigate the performance improvement of introducing texture super-resolution in the explicit 2D texture mapping space. \refFig{fig:ablation_sr} shows introducing super-resolution textures reduces noise in the images and restores more details. In addition, we try to optimize neural implicit representation directly under the supervision of super-resolution images. However, \refFig{fig:ablation_sr} shows that it will blur local details (such as the buttons). This is because the corresponding projection within the deformable field from pose space to canonical space is not stable, which results in the multiple points with different colors in pose space projected to a single point in canonical space and further results in the blurring of textures. We perform super-resolution in the explicit texture space, which ensures a stable correspondence between geometry and color and produces more clearer rendering result as shown in \refTab{tab:ablation_unbias and sr}.

\section{Conclusion}
\label{sec:Limitation and Conclusion}
This paper proposes HR human, a novel framework that enables reconstructing a digital avatar equipped with triangular mesh and corresponding PBR material texture from a monocular camera. We introduce a novel information fusion strategy to combine the information from the monocular video and synthesize virtual multi-view images to compensate for the missing spatial view information. In addition, we correct the bias for the boundary and size of the mesh extracted from the implicit field. Finally, we introduce a pretrained latent diffusion model to distill the super-resolution texture when jointly optimizing the mesh and texture. The high-quality mesh and high-resolution texture produced by our method are compatible with common modern engines and 3D tools, which simplify the modeling process of digital avatars in various downstream applications and can be directly edited and reilluminated.

\bibliographystyle{ACM-Reference-Format}
\bibliography{sample-base}

\clearpage
\appendix

\section{Training Strategy}

We divide the complete training into two stages. In the first stage, we trained the deformable implicit neural representation based on volume rendering. We typically took 100k iterations for the first stage. Then, we use marching cubes to extract a mesh in a well-behaved pose from deformable implicit neural fields. Further, we take 15k iterations for the second stage of training. In the second stage, we aim to optimize the PBR material textures, lighting, and triangular mesh using a differentiable PBR-based render layer. Specifically, we first apply an unbiased optimization to adjust the coarse mesh extracted from the first stage($\sim$ 1k iters), resulting in a finetuned mesh aligned with the real human. Equipped with the finetuned mesh, we optimized the corresponding PBR texture under the supervision of sparse real camera views and dense synthesis views. After the coarse texture had converged($\sim$ 10k iters), we adapted prior knowledge of the latent diffusion model at super-resolution in multi-view rendering to distill the texture. In practice, the Adam optimizer was employed for optimizing all networks and parameters. We set the learning rate to $5\times10^{-4}$ with an exponential falloff during the optimization. The entire experiment was trained on an NVIDIA A100 GPU.
\label{sec:training details}

\section{Loss Function}

The definition of the loss functions mentioned in the main paper for the training of the first stage includes,
$\mathcal{L}_{color}$ follows the $L1$ loss:
\begin{equation}
\mathcal{L}_{color}=\sum_{r\in R}\left \| \hat{C}_{i}(r)-C_{i}(r) \right \|_{1}.
\label{eq:colorl1loss}
\end{equation}

$\mathcal{L}_{eik}$ is the Eikonal term~\cite{eikonalloss} encouraging $f_s$ to approximate a signed distance function, and we set $\lambda_1$ as $0.1$:
\begin{equation}
\mathcal{L}_{eik}=\lambda_1 \sum_{x'}(\left \| \nabla f_s(x')\right \|_{2}-1)^2,
\end{equation}

$\mathcal{L}_{curv}$ is the curvature term~\cite{permutosdf} encouraging to recover smoother surfaces in reflective or untextured areas:
\begin{equation}
\mathcal{L}_{curv}=\lambda_2 \sum_{x'}(n \cdot n_{\epsilon}-1)^2,  
\end{equation}
where $n = \nabla f_s(x')$ are the normal at the points $x'$, $n_\epsilon = \nabla f_s(x'_\epsilon)$ are the normal at perturbed points $x'_\epsilon$.The perturbed points $x'_\epsilon$ are sampled randomly in tangent plane, $x'_\epsilon = x' + \epsilon (n \times \tau)$. $\tau$ is random unit vector. And we set $\lambda_2$ as $0.65$. 

$\mathcal{L}_{offset}$ is the regularization term, which constrains the non-rigid deformation within a small range. We set $\lambda_3$ as 0.02 and the loss is defined as: 
\begin{equation}
\mathcal{L}_{offset} = \lambda_3 \overline {\left \| \Delta x\right \|}_{2}
\end{equation}

In addition, we use a consistency regularization term $\mathcal{L}_{w}$ to minimize the difference between blend weights of the canonical and observation spaces which are supposed to be the same. The loss is defined as:
\begin{equation}
\mathcal{L}_{w}=\sum_{x}\left \| w_{i}(x) - w^{can}_i (x') \right \|_{1},
\end{equation}

The definition of the loss functions mentioned in the main paper for the training of the second stage includes,
$\mathcal{L}_{render}$ also follows the $L1$ loss:
\begin{equation}
\mathcal{L}_{render}=\sum_{r \in R}\left \| \hat{C}_{i}(x_s)-C_i(r) \right \|_{1}.
\end{equation}

$\mathcal{L}_{mask}$ is applied in early epochs(such as 10), to estimate the residual of biased surface. It is defined as:
\begin{equation}
\mathcal{L}_{mask} = \sum_{r \in R}\left \| \hat{M}_{i}(x_s)-M_i(r) \right \|_{2}.
\end{equation}
where $\hat{M}_{i}(x_s)$ is the mask after rasterization.

$\mathcal{L}_{light}$ is regularization term~\cite{nvdiffrec} designed to penalizes color shifts. $\lambda_4$ is set as 0.005. Given the per-channel average intensities $\hat{c_i}$, we define it as:
\begin{equation}
\begin{aligned} L_{light} &= \lambda_4 \frac{1}{3}\sum_{i=0}^{3}\left | \hat{c_i} - \frac{1}{3} \sum_{i=0}^{3} \hat{c_i} \right |
\end{aligned}
\end{equation}

$\mathcal{L}_{smooth}$ is a smooth term that calculate texture differences between surface points $x_s$ and its random displacement $x_s+\epsilon$. $\lambda_5$ is set as 0.002. And we define it as:
\begin{equation}
\mathcal{L}_{smooth} = \lambda_5 \sum_{x_s} \left | k_d(x_s)-k_d(x_s+\epsilon) \right |   
\end{equation}

$\mathcal{L}_{SDS}$ is defined as eq.15 of the main paper and is activated after the $\mathcal{L}_{render}$ converges.

\section{More Results and Application}
\label{sec:more_results}
\begin{figure*}[!h] 
    \centering 
    \includegraphics[width=\linewidth]{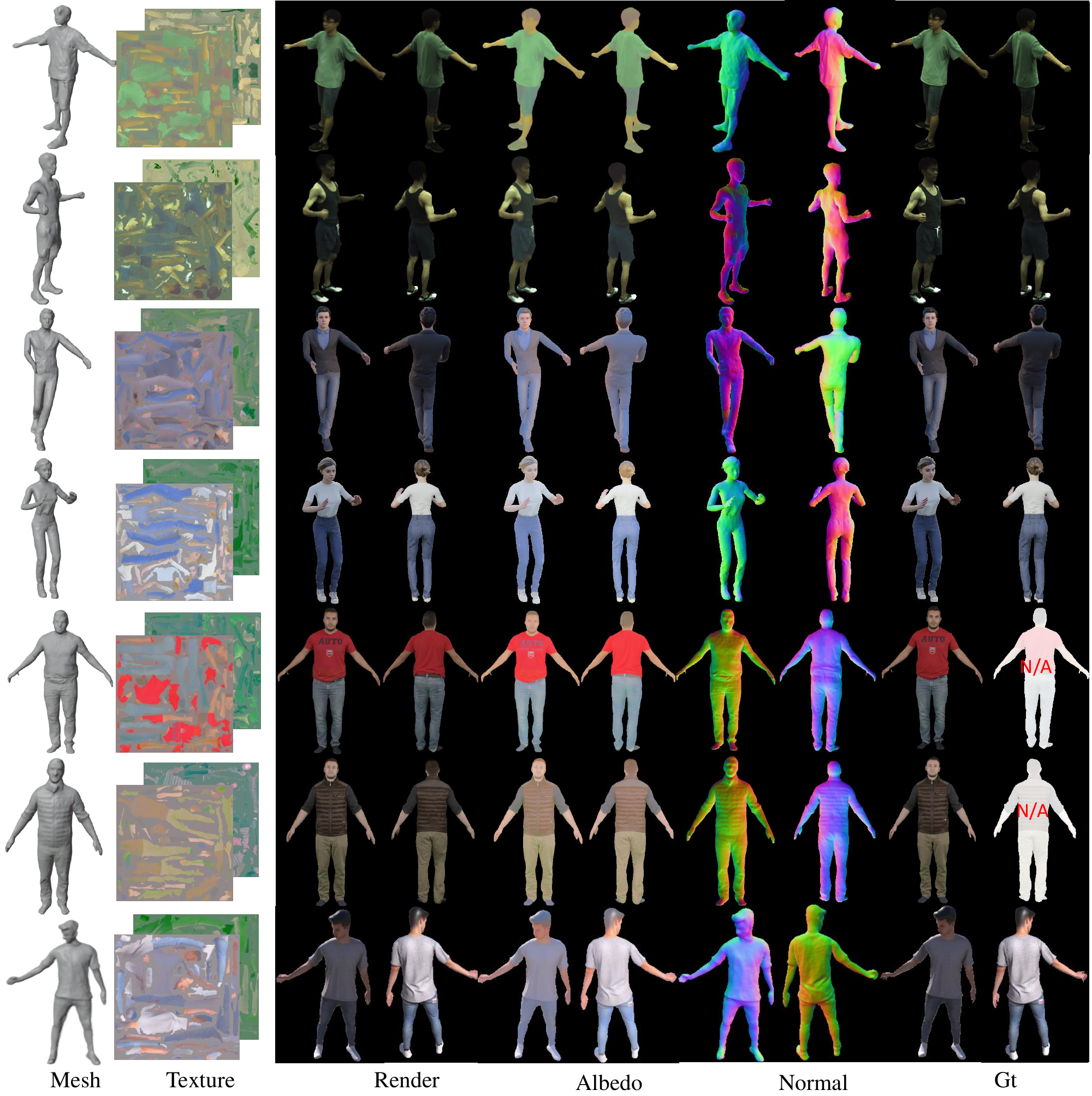}
    \caption{More high-quality reconstruction results of human body geometry and texture materials, including real-world dataset "ZJU-MoCap" and "PeopleSnapshot", synthesized dataset "Renderpeople". From top to bottom, they are "CoreView313", "CoreView377", "Josh", "Megan", "M3C", "M2C", "Manuel"} 
    \label{fig:more_result} 
\end{figure*}

\begin{figure*}[!h] 
    \centering 
    \includegraphics[width=\linewidth]{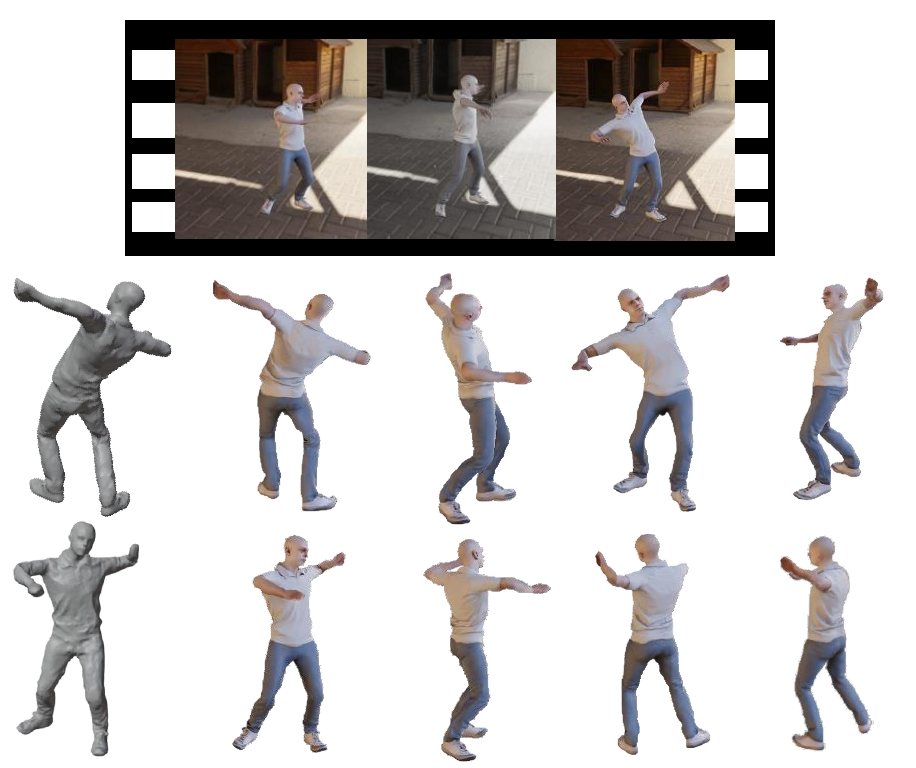}
    \caption{Reconstruction results of different frames in a dance video of dataset "RenderPeople-Brain".} 
    \label{fig:brain} 
\end{figure*}

\begin{figure*}[!h] 
    \centering 
    \includegraphics[width=\linewidth]{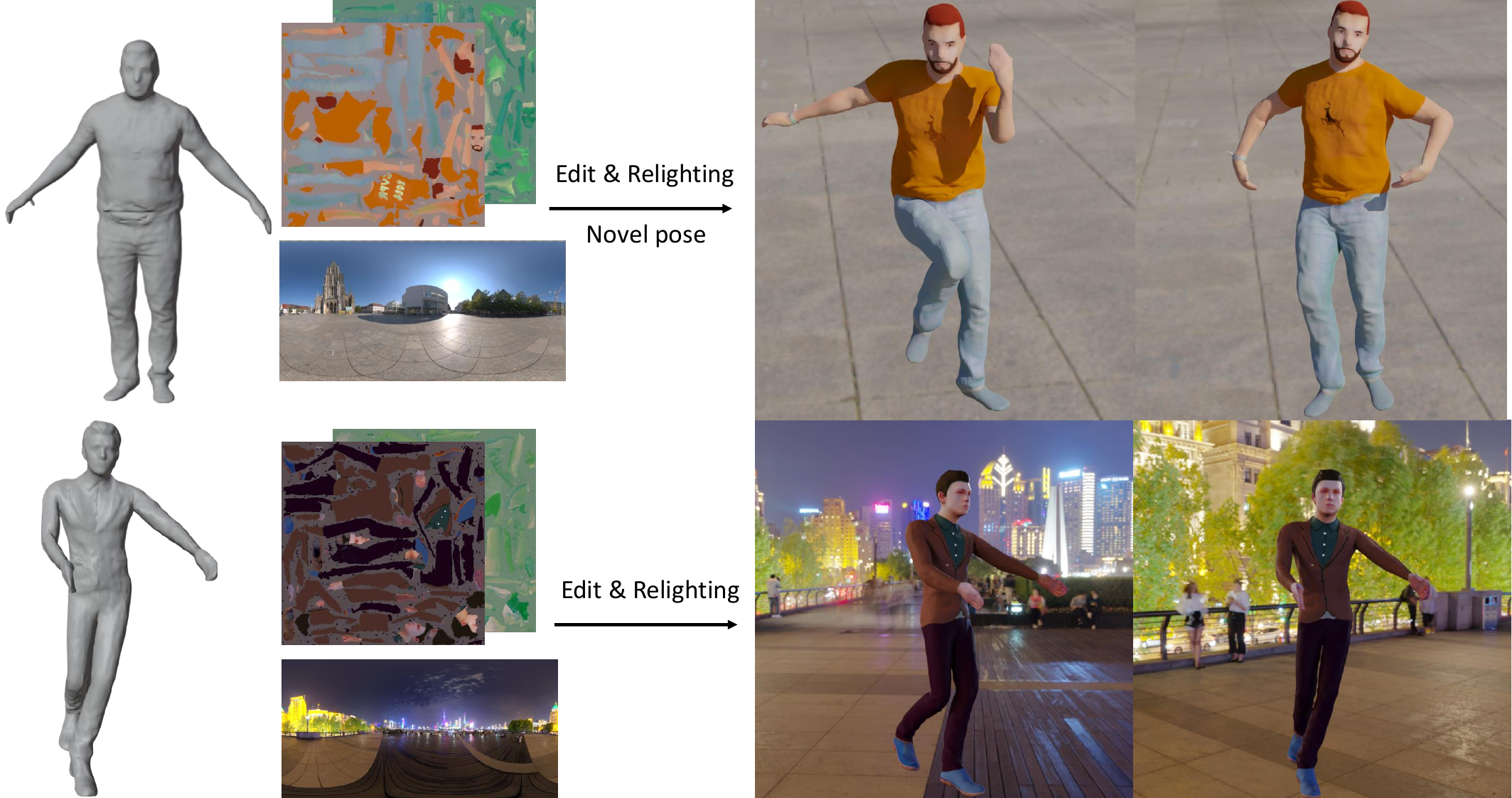}
    \caption{The rendering result of texture edit,relighting and novel poses synthesis, that work on graphics engines.} 
    \label{fig:edit_result} 
\end{figure*}

We present full results for both synthesized datasets and real-world datasets in \refFig{fig:more_result} and \refFig{fig:brain}, including meshes, texture materials, and rendering results. Meanwhile, to demonstrate the compatibility of our results in standard graphics engines, \refFig{fig:edit_result} shows the results of \textbf{relighting}, \textbf{texture editing}, \textbf{novel poses synthesis} on "M3C" and "Josh" dataset. 

\section{Additional Experiments}
\begin{figure}[!h] 
    \centering 
    \includegraphics[width=\linewidth]{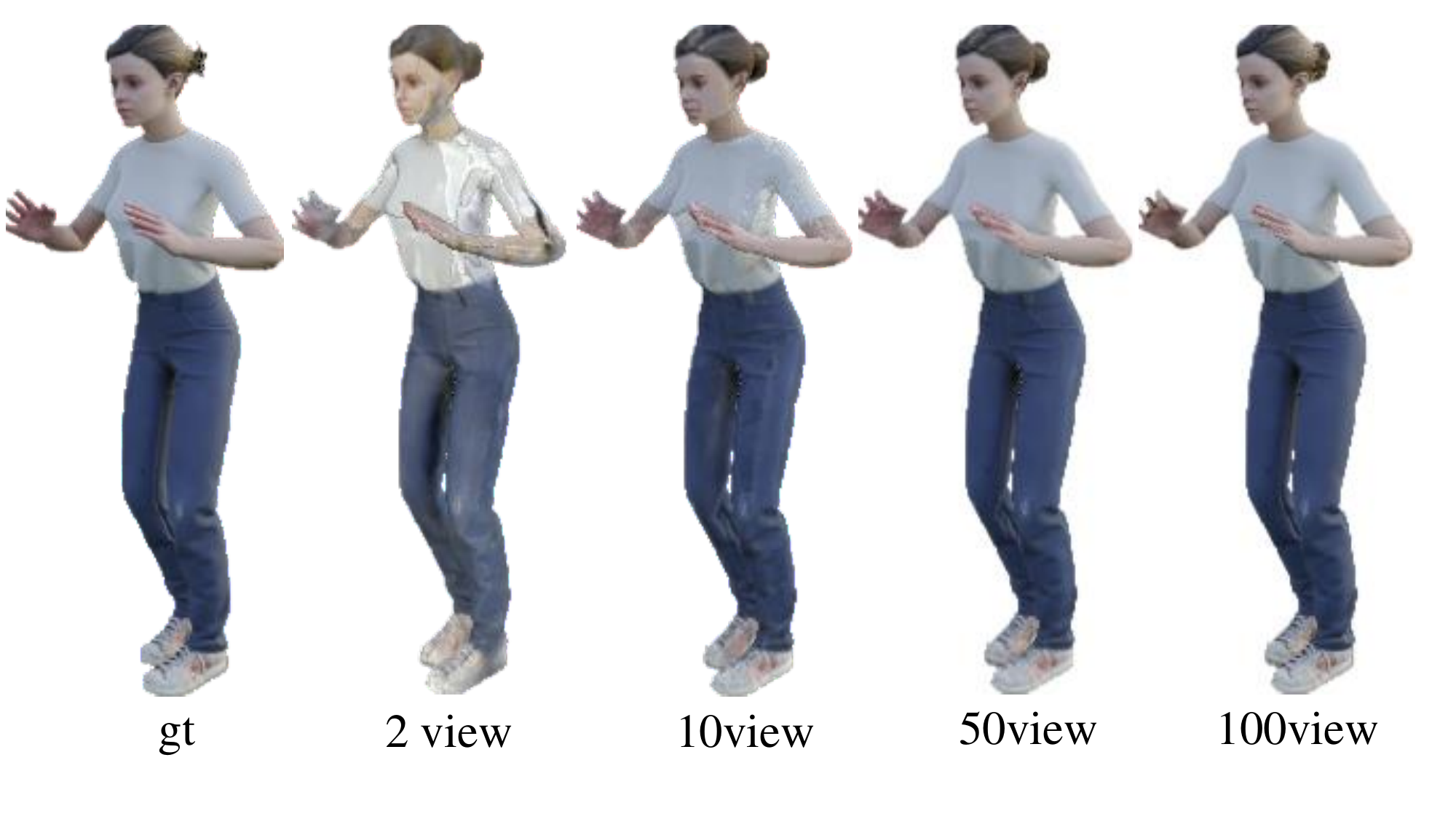}
    \caption{Qualitative comparison of the effectiveness of the number of synthesized views used in fusion strategy. From left to right, the number of training views is increasing.} 
    \label{fig:ablation_viewfig} 
\end{figure}

\textbf{\emph{The effectiveness of the number of synthetic views used in information fusion strategy.}} We designed four controlled to find the most suitable number of synthetic views for the fusion training. \refFig{fig:ablation_viewfig} shows the comparison of training results from different numbers of synthetic views in novel view. \refTab{tab:ablation_view} shows that increasing the number of synthesized views is beneficial for learning more detailed textures because unknown surfaces are reduced. We usually choose 50 views as the training baseline to balance training efficiency and effectiveness.

\begin{table}[!h]
\centering
\setlength{\tabcolsep}{5.4mm}
\begin{tabular}{cccc}
\hline
VIEWS  & PSNR  & SSIM  & LPIPS \\ \hline
2   & 19.71 & 0.794 & 0.136 \\
10  & 22.54 & 0.853 & 0.11  \\
50  & 23.13 & 0.91  & 0.057 \\
100 & 23.27 & 0.92  & 0.045 \\ \hline
\end{tabular}
\caption{Quantitative comparison for the effectiveness of the number of synthesized views used in fusion strategy on “Megan” dataset.}
\label{tab:ablation_view}
\end{table}

\begin{figure}[!h] 
    \centering 
    \includegraphics[width=\linewidth]{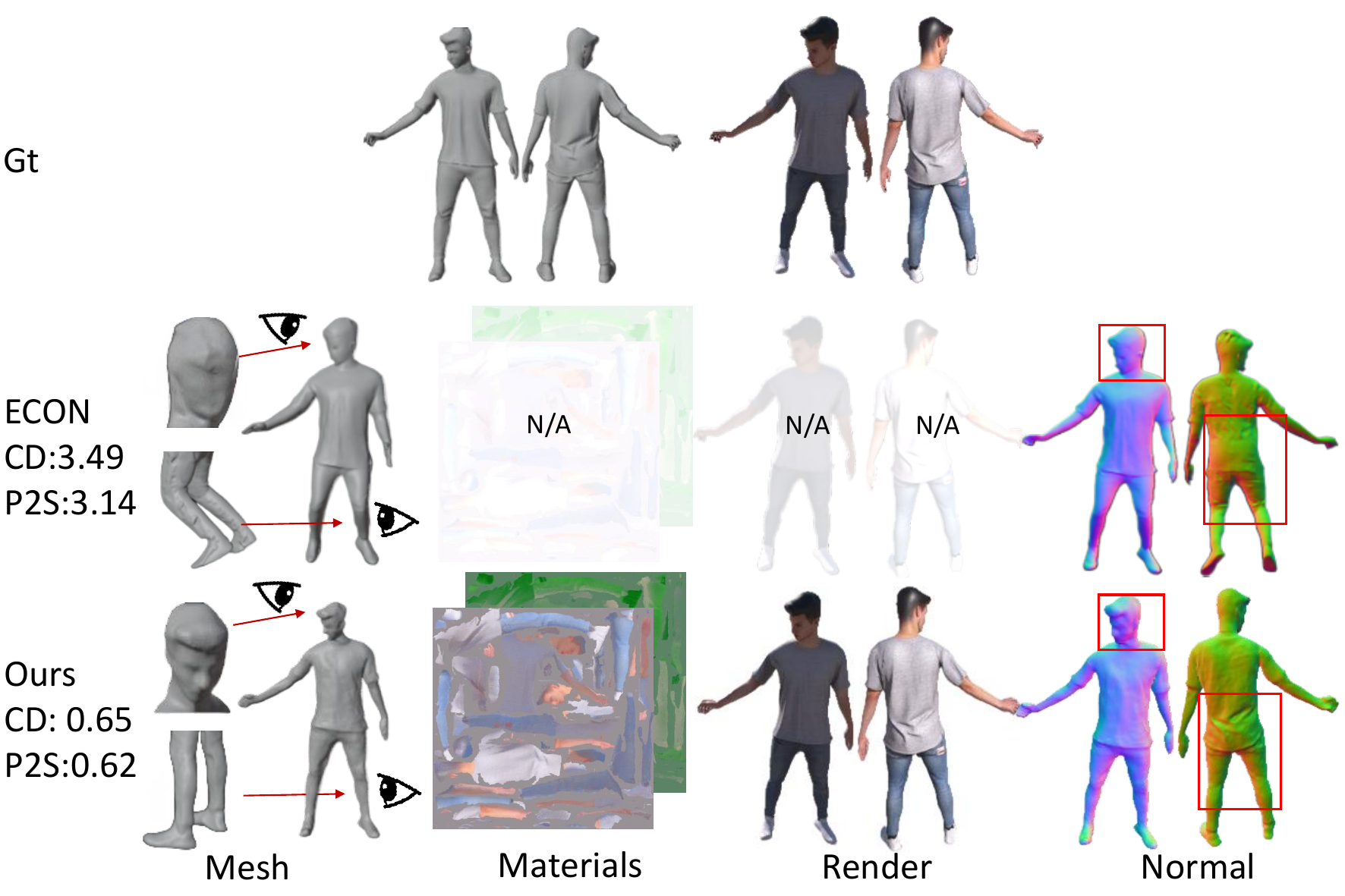}
    \caption{Comparison result on "RenderPeople" dataset.} 
    \label{fig:econ} 
\end{figure}

\textbf{\emph{Difference with single-image based explicit human reconstruction.}} We designed a comparative experiment with the SOTA method of the single-image based model, ECON\cite{econ}. ECON is a single-image based explicit human reconstruction method that combines implicit representation and explicit body regularization. Unlike ECON, which only reconstructs geometry, our approach further delivers triangular meshes and PBR textures, both of which are highly valued as 3D assets in the industry. As shown in \refFig{fig:econ}. Our method offers accurate full-body geometry, including details of the face, back, and legs.

\begin{figure}[!h] 
    \centering 
    \includegraphics[width=\linewidth]{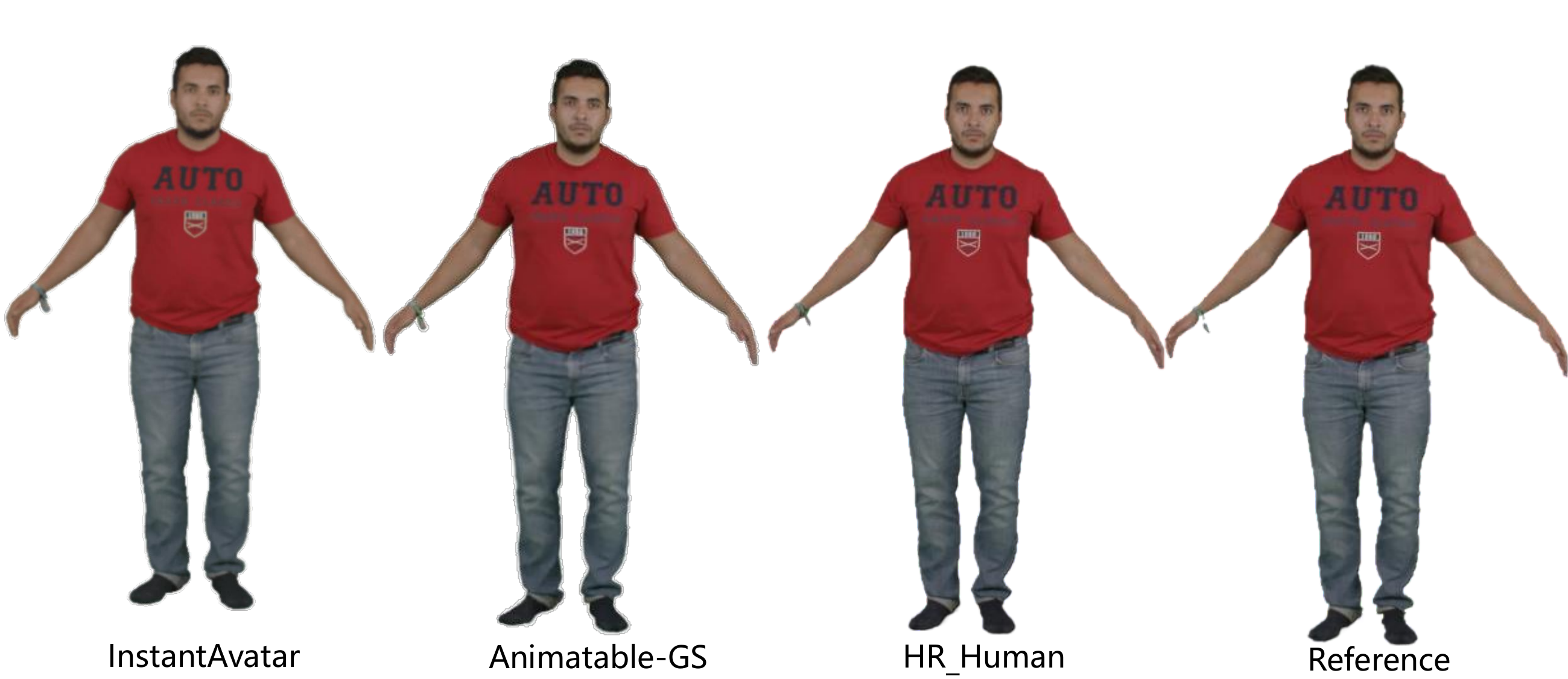}
    \caption{Comparison result on "PeopleSnapshot" dataset.} 
    \label{fig:gs_exp} 
\end{figure}

\textbf{\emph{Difference with SOTA of real-time neural rendering}} We designed a comparative experiment with InstantAvatar\cite{instantavatar} and Ani-3DGS\cite{ani-3dgs}. Both of these are based on surface point priors and achieve real-time performance. But they all ignore the decoupling of PBR materials and high-precision geometry. Because of deformation residual field, our approach does not rely on surface initialization and pose. Thus, reflected in the metrics, our approach have more accurate surface and multi-view consistent textures in real-world data. Meanwhile, our approach further delivers triangular meshes and PBR textures, which supports for direct editing and relighting in common graphics engine. Although a unityGS\cite{unitygs} compatibility plugin has been released in the community, the obvious software limitations and lack of editing and relighting are truly concerning. And as shown in \refFig{fig:gs_exp} and \refTab{tab:gs_exp}, we have to admit that we have overlooked the importance of train cost, but getting clearer, editable and relightable textures and geometry are worth it. We will prioritize real-time performance in future work, such as using CUDA acceleration. We hope to inspire more researchers and engineers to consider editability, relighting and compatibility.
\begin{table}[!h]
\centering
\begin{tabular}{ccccc}
\hline
  & PSNR  & SSIM  & LPIPS & Train Cost \\ \hline
InstantAvatar   & 26.61 & 0.93 & 0.12 & <5min \\
Animatable GS  & 29.81 & 0.974 & 0.023 & <1h \\
Ours  &32.4 & 0.971  & 0.017 & >1h \\ \hline
\end{tabular}
\caption{Quantitative comparison of different rendering methods on “PeopleSnapshot” dataset.}
\label{tab:gs_exp}
\end{table}

\section{Architecture}
\label{sec:architecture}
\begin{figure*}[!h] 
    \centering 
    \includegraphics[width=\linewidth]{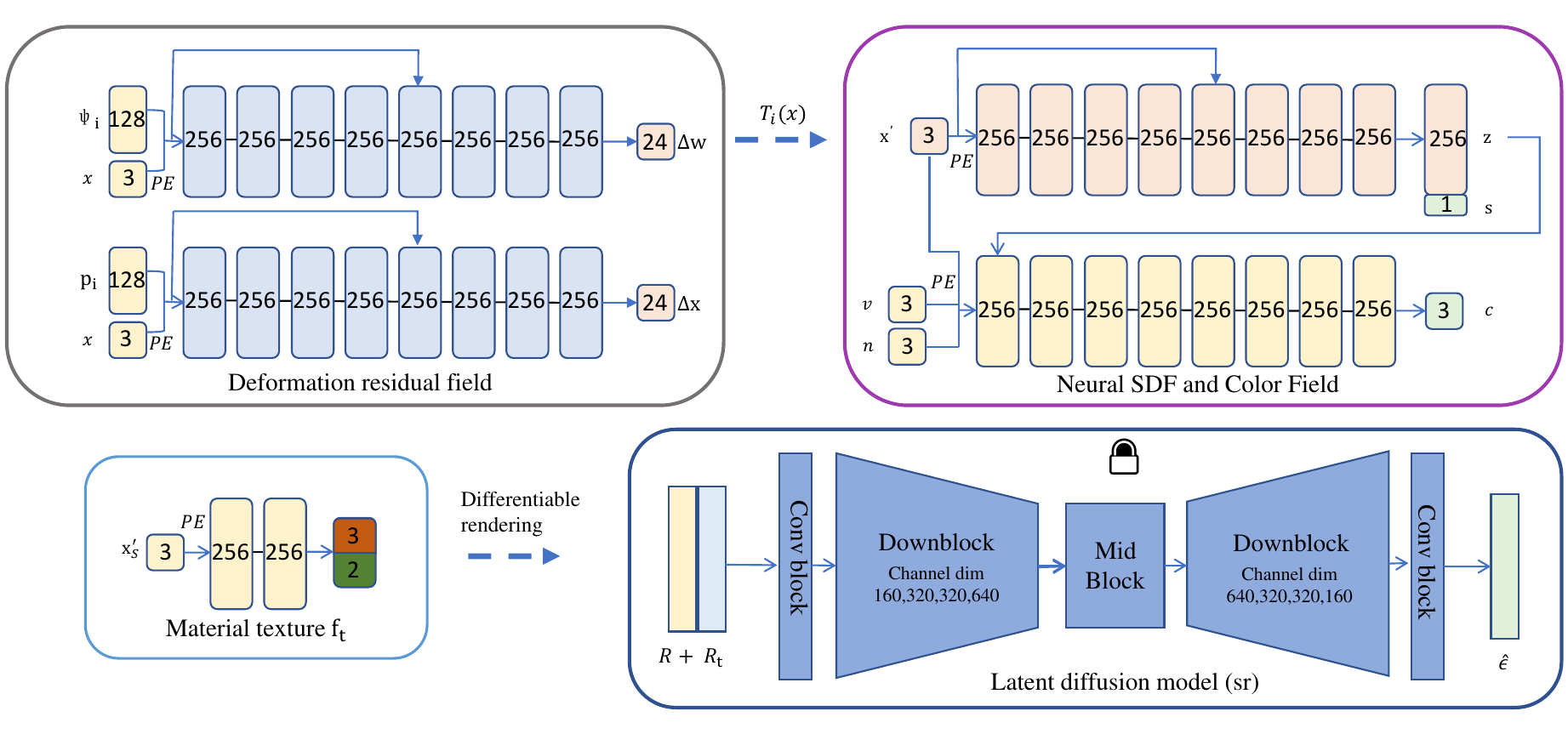}
    \caption{Visualization of the architecture of the proposed framework, where PE represents position encoding} 
    \label{fig:architecture} 
\end{figure*}

As shown in \refFig{fig:architecture}, our method consists of five MLP networks and a latent diffusion model. We give the number of layers and dimensions for each network. In addition, the super-resolution model uses the latent diffusion pre-trained model trained by\cite{ldm}. \textbf{Additionally, we will make our code and dataset available to the community after paper acceptance, facilitating testing on more datasets}.

\section{Limitation}
\label{sec:training details}
Our method still has the following limitations. Firstly, we model the human without distinguishing clothes and the human body. Thus, our method does not apply to humans wearing complex or loose clothing. In addition, our method still lacks competitiveness in terms of training costs, we will consider introducing  CUDA acceleration or other strategy. Finally, we choose to accelerate optimization without considering global lighting. Therefore, there is still room for improvement in the decoupling of our materials.

\label{sec:architecture}

\end{document}